\documentclass[sigconf]{acmart}
\usepackage{amsmath}
\usepackage{multirow}

\AtBeginDocument{%
  \providecommand\BibTeX{{%
    \normalfont B\kern-0.5em{\scshape i\kern-0.25em b}\kern-0.8em\TeX}}}

\copyrightyear{2023}
\acmYear{2023}
\setcopyright{acmlicensed}
\acmConference[MM '23] {Proceedings of the 31st ACM International Conference on Multimedia}{October 29--November 3, 2023}{Ottawa, ON, Canada.}
\acmBooktitle{Proceedings of the 31st ACM International Conference on Multimedia (MM '23), October 29--November 3, 2023, Ottawa, ON, Canada}
\acmPrice{15.00}
\acmISBN{979-8-4007-0108-5/23/10}
\acmDOI{10.1145/XXXXXX.XXXXXX}

\acmSubmissionID{3661}

\begin{document}
\title{Contrast-augmented Diffusion Model with Fine-grained Sequence Alignment for Markup-to-Image Generation}

\author{Guojin Zhong}
\affiliation{%
  \institution{College of Computer Science and Electronic Engineering, Hunan University}
  \city{Changsha}
  \country{China}}
\email{gjzhong@hun.edu.cn}

\author{Jin Yuan}
\authornote{Corresponding author.}
\affiliation{%
  \institution{College of Computer Science and Electronic Engineering, Hunan University}
  \city{Changsha}
  \country{China}}
\email{yuanjin@hnu.edu.cn}

\author{Pan Wang}
\affiliation{%
  \institution{College of Computer Science and Electronic Engineering, Hunan University}
  \city{Changsha}
  \country{China}}
\email{wpanda@hun.edu.cn}

\author{Kailun Yang}
\affiliation{%
 \institution{School of Robotics, Hunan University}
 \city{Changsha}
 \country{China}}
\email{kailun.yang@hnu.edu.cn}

\author{Weili Guan}
\affiliation{%
  \institution{Monash University}
  \city{Melbourne}
  \country{Australia}}
\email{honeyguan@gmail.com}

\author{Zhiyong Li}
\authornotemark[1]
\affiliation{%
  \institution{School of Robotics, Hunan University}
  \city{Changsha}
  \country{China}}
\email{zhiyong.li@hnu.edu.cn}

\renewcommand{\shortauthors}{G. Zhong, J. Yuan, P. Wang and K. Yang, et al.}

\begin{abstract}
The recently rising markup-to-image generation poses greater challenges as compared to natural image generation, due to its low tolerance for errors as well as the complex sequence and context correlations between markup and rendered image. This paper proposes a novel model named ``Contrast-augmented Diffusion Model with Fine-grained Sequence Alignment'' (FSA-CDM), which introduces contrastive positive/negative samples into the diffusion model to boost performance for markup-to-image generation. Technically, we design a fine-grained cross-modal alignment module to well explore the sequence similarity between the two modalities for learning robust feature representations. To improve the generalization ability, we propose a contrast-augmented diffusion model to explicitly explore positive and negative samples by maximizing a novel contrastive variational objective, which is mathematically inferred to provide a tighter bound for the model's optimization. Moreover, the context-aware cross attention module is developed to capture the contextual information within markup language during the denoising process, yielding better noise prediction results. Extensive experiments are conducted on four benchmark datasets from different domains, and the experimental results demonstrate the effectiveness of the proposed components in FSA-CDM, significantly exceeding state-of-the-art performance by about $2\% \sim 12\%$ DTW improvements. The code will be released at https://github.com/zgj77/FSACDM.
\end{abstract}



\maketitle

\section{Introduction}
\begin{figure*}
    \centering
    \includegraphics[width=0.77\textwidth]{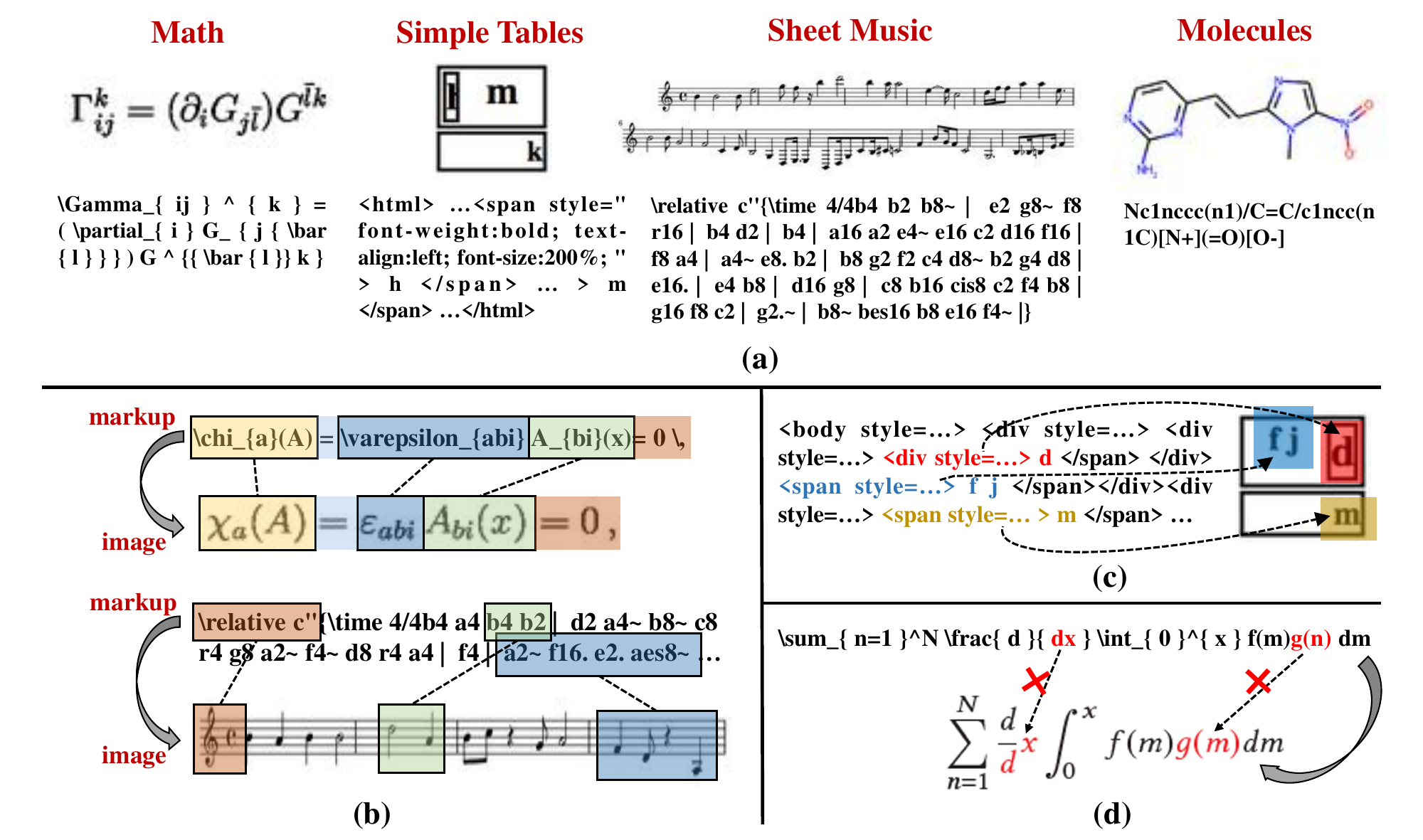}
    \caption{Several examples to illustrate the datasets of markup-to-image (Sub-figure (a)) as well as its characteristics, including the sequence relationship between markups and images (Sub-figure (b)), the contextual correlation within a markup description (Sub-figure (c)), and the low tolerance for character errors (Sub-figure (d)).}
    \label{fig:challenge statement}
\end{figure*}
The recent progress of generative models like Generative Adversarial Network (GAN) \cite{Wu_Adma-gan22, Wang_CycleGan21} and Denoising Diffusion Probabilistic Model (DDPM) \cite{Jonathan_nips_2020} has tremendously promoted the prosperity of text-to-image generation, which aims at generating a semantically matching image conditioned on a language description. Most existing image generation studies focus on generating natural images, where each ambiguous natural language expression may generate multiple semantically consistent images with diverse appearances.

As a comparison, Deng \textit{et al.} \cite{deng2022markup} recently proposed a novel text-to-image task called ``markup-to-image'', which aims at mapping a structured markup description like mathematical formulas, HTML simple tables, music notations, and chemical molecules (see Figure \ref{fig:challenge statement} (a)), into an exact image, which could precisely express the corresponding markup language. Different from natural image generation, markup-to-image task has known ground truth to facilitate models' evaluation. Moreover, the compositional nature of markup language requires a deeper exploration of relational properties, which poses greater challenges for this task.

Benefiting from advances in Diffusion Models (DMs) for image generation \cite{9878449}, Deng \textit{et al.} first designed a DM with scheduled sampling for markup-to-image generation. This approach attempts to alleviate the exposure bias problem in DMs, where a model is never exposed to incorrectly generated tokens during the training but frequently faces these errors in the inference phase. Despite the success, it still has the following drawbacks when applied in markup-to-image generation: First, it does not explore the fine-grained temporal alignment relationship between a markup language description and the corresponding rendered image, which is widely present in the data of this task (see Figure \ref{fig:challenge statement} (b)); Second, there is strong compositional nature of markup language, and capturing the contextual correlation within a markup description is crucial for a deep understanding (see Figure \ref{fig:challenge statement} (c)); Third, markup-to-image has a low tolerance for errors, and even a small symbol error may cause a complete semantic bias (see Figure \ref{fig:challenge statement} (d)). Thus, using a single image construction process in DMs without positive/negative contrastive feedback is not conducive to improving the model's generalization.

Towards this end, this paper proposes a novel \textbf{C}ontrast-augmented \textbf{D}iffusion \textbf{M}odel with \textbf{F}ine-grained \textbf{S}equence \textbf{A}lignment (FSA-CDM), which exposes the model to both positive and negative samples with powerful feature learning for markup-to-image generation. Specifically, given a pair of image and markup inputs, we first encode them into a sequence of visual and language tokens, respectively. Then, we employ Bi-LSTM to capture the contextual relationship among visual tokens and design a fine-grained cross-modal alignment module to well align each visual token with the corresponding textual token sequence by sequence. On this basis, we propose a contrast-augmented diffusion model for markup image generation. Different from \cite{deng2022markup} only receiving a single sample for construction, FSA-CDM receives a to-be-constructed sample with several contrastive positive/negative samples to augment the model's training. We mathematically design a contrastive variational objective integrating both positive and negative samples and infer a tighter bound for optimization. By contrastively exploring multiple samples, FSA-CDM could better improve the model's generalization ability as well as reduce prediction errors. Furthermore, we design a Context-aware Cross Attention Module (CCAM) to replace the traditional cross attention during the denoising process. CCAM constructs a relationship matrix from visual features to explore the complex contextual relationship among markup inputs, and thus could better predict noise for image construction.   

Extensive experiments are carried out on all four benchmark datasets from different domains. Our FSA-CDM presents high-quality image generation and significantly outperforms state-of-the-art (SOTA) methods on all the benchmark datasets. Our contributions are summarized as follows:
\begin{itemize}
    \item We propose a fine-grained sequence alignment module to align markup language and its rendered image at the sequence level, thereby learning robust uni-modal representations to support markup-to-image generation.
    \item We propose a contrast-augmented diffusion model that explicitly introduces positive and negative samples by using contrastive learning. A novel contrastive variational objective is mathematically inferred to achieve a tighter bound for variational evidence, improving the model's generalization ability as well as alleviating the exposure bias problem. 
    \item We design a context-aware cross attention module for noise prediction during the denoising process. CCAM constructs a relationship matrix between characters from visual features to guide the contextual information capturing among markup language, yielding accurate noise prediction. 
\end{itemize}

\section{Related Work}
\subsection{Text-to-Image Generation}
The early text-to-image generation adopts Generative Adversarial Networks (GAN) \cite{goodfellow2020generative}, which was first proposed by Reed \textit{et al.} \cite{2016arXiv160505396R}. On this basis, a variety of studies on GANs have been proposed to improve the quality of image generation via progressive refinement \cite{Zhang_2017_iccv,stackGAN}, cross-modal attention \cite{Attngan, Zhanghan_2021_cvpr, Shi_AtHom22} as well as semantic modeling \cite{Qiao_RGan21, Chen_background22,df-gan}. Another text-to-image generation paradigm adopts VQ-VAE \cite{vqvae,vqvae2} to generate discrete image markers from text cues. Imposing VQ-VAE, transformer-based approaches like DALLE \cite{DALLE} and CogView \cite{CogView} could effectively generate images from text prompts but suffer from the limitations of autoregressive models with one-way bias and cumulative prediction errors \cite{Radford2019LanguageMA}. 

Diffusion model (DM) \cite{Jonathan_nips_2020, GLIDE} is a recently rising approach in text-to-image generation, which attempts to add noises into an image step by step followed by the denoising process to reconstruct the image. Compared to GAN-based methods, DMs are free of training instability and mode collapse \cite{beatgan}, and thus demonstrate impressive performance for image generation \cite{imagen, DALL-E2}. For instance, Gu \textit{et al.} proposed VQ-Diffusion \cite{Gu_VQ-diffusion} based on VQ-VAE to eliminate the one-way bias and cumulative prediction errors, yielding better image quality. Liu \textit{et al.} \cite{Liu2022CompositionalVG} proposed a composable DM to solve the semantic deficiency, where an image is generated by a set of diffusion models, each modeling a component of the image. Benny \textit{et al.}~\cite{benny2022dynamic} proposed a dynamic dual-output DM to address the problem of poor-quality images when the number of iterations is low. Besides, several studies aimed at improving the understanding of text inputs. For instance, Xu \textit{et al.} \cite{xu2022odise} proposed a method combining CLIP, which uses high-quality images generated by a pre-trained text DM for all-view segmentation.
Zhao \textit{et al.} further proposed MagicFusion \cite{zhao2023magicfusion} to fuse multiple text-guided DMs to improve image quality. Gao \textit{et al.} proposed a Masked Diffusion Transformer \cite{gao2023masked}, which can reconstruct the complete information of an image from an incomplete contextual input. More advanced researches on DMs for image generation include DMs on semi-supervised learning~\cite{zhou2023shifted}, DMs on attention mechanism \cite{Hila_Ateend_excite}, \textit{etc.} Recently, Deng~\textit{et al.}~\cite{deng2022markup} first proposed a novel image generation task named markup-to-image by DM. Different from natural image generation with flexible interpretations of text prompts, markup-to-image generation aims to generate an exact image under a unique interpretation of a given markup prompt and has a low tolerance for symbol errors, which greatly increases the technical challenges. 

\subsection{Contrastive Learning in Generative Model}
Contrastive learning is a powerful self-supervised representation learning \cite{Chen_represent} approach that has been used in generative models. For instance, Kang \textit{et al.} proposed ContraGAN \cite{Kang_gancl2} to consider the relationship of multiple image embeddings as well as the data-to-class relationship by using a conditional contrast loss. Yang \textit{et al.} proposed DiscoFaceGAN \cite{Yang_ganl3} to add contrastive learning to face generation to facilitate untangling, allowing precise control of face attributes. Ye \textit{et al.} \cite{Ye_imporving} proposed a contrastive learning method to learn consistent textual representations of captions corresponding to the same image, thereby enhancing the quality and semantic consistency of synthetic images. In addition, Parmar \textit{et al.} indicated that contrastive learning can be combined with metric learning to improve VAE, solving the instance-level fidelity between input and reconstruction in the induced feature space. ContrastVAE \cite{wang2022contrastvae} creates two views for input data to alleviate the uncertainty and sparsity issues, thereby improving the generalization ability of VAE. For diffusion models, Ouyang \textit{et al.} \cite{ouyang2022improving} found that enhancing distinguishability was important and thus adopted the contrastive loss to guide the diffusion model. In order to improve the connection between input and output, Zhu \textit{et al.} \cite{zhudiscrete} designed a conditional discrete contrastive diffusion loss, which directly incorporates negative samples into the model's training for optimizing the evidence lower bound. Although the above methods have achieved good results in image generation, they simply introduce the contrastive loss into DMs in a separate learning pattern or only incorporate negative samples to DMs to optimize conventional variational objectives, which has limited learning capability on image generation.

Beyond the previous studies, this work explicitly incorporates both positive and negative samples into DM for better optimizing variational objectives to provide a tighter bound for evidence, thereby improving discrimination and generalization performance for markup-to-image generation. Moreover, our approach first explores sequence and context relationships for markup-to-image generation, and thus could better capture the correlations between visual and textual features for performance improvement.  

\section{Method}
\begin{figure*}
    \centering
    \includegraphics[width=0.78\textwidth]{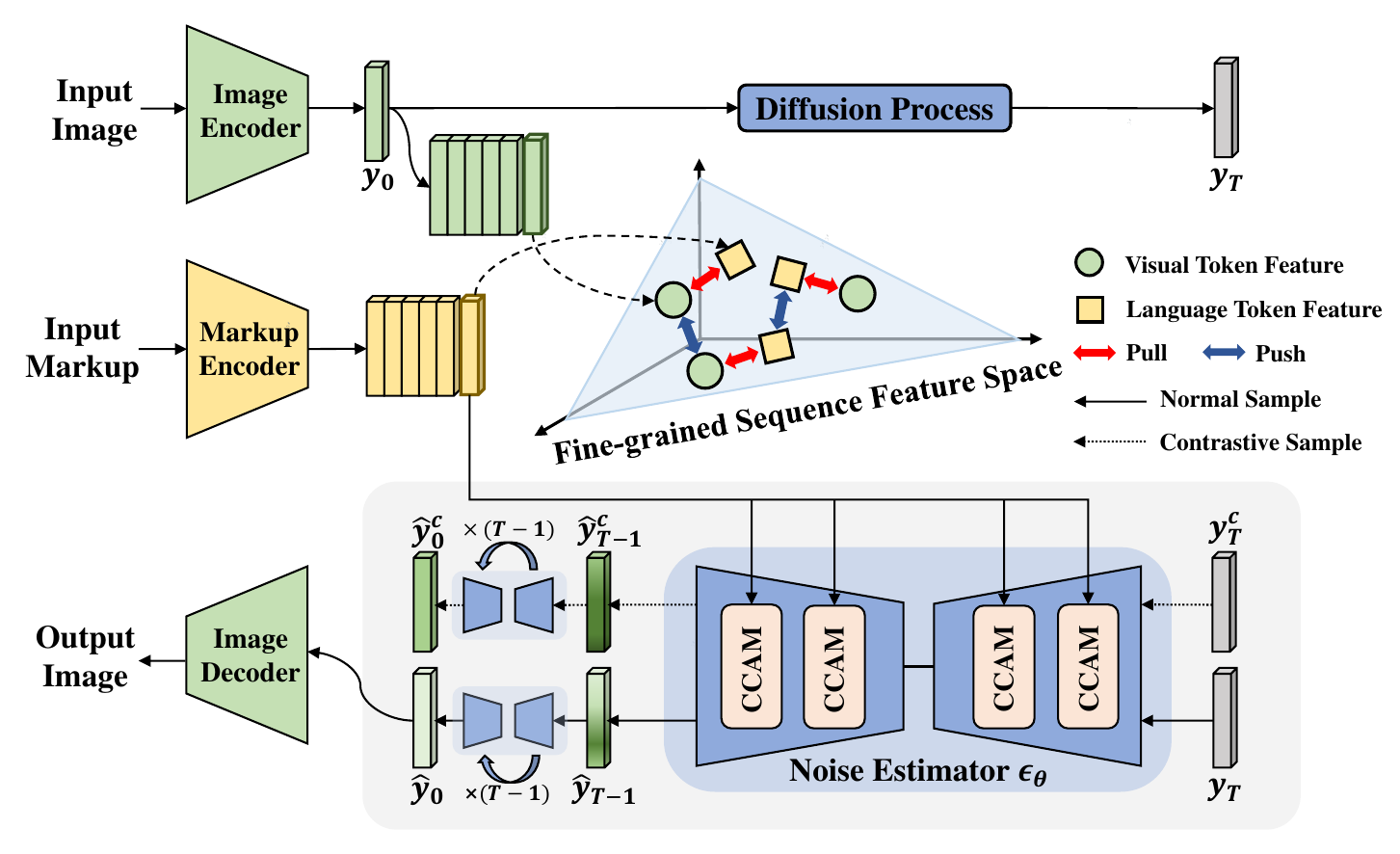}
    \caption{The framework of FSA-CDM, which consists of an image and a markup encoder with fine-grained cross-modal alignment, and a contrast-augmented diffusion model containing several CCAMs, where $y_0$, $y_T$, $\hat{y}_t$ and $y^c_t$ denote original samples, full-noise samples, denoising samples, and contrast-augmented positive/negative samples, respectively.}
    \label{fig:Framework}
\end{figure*}

In this section, we first present the definition of markup-to-image generation, and then describe our Contrast-augmented Diffusion Model with Fine-grained Sequence Alignment (FSA-CDM). The overall framework of FSA-CDM is illustrated in Figure \ref{fig:Framework}, which consists of two parts: an image and a markup encoder with fine-grained sequence alignment to extract robust uni-modal representations, and a contrast-augmented diffusion model with a context-aware attention module to accurately generate rendered images.

\subsection{Task Definition}
Given a markup language $x \in \mathcal{X}$ and its corresponding rendered image $y \in \mathcal{Y}$, the goal of markup-to-image is to establish a model $f_{\theta}: \mathcal{X} \rightarrow \mathcal{Y}$ to approximate the true mapping $f: \mathcal{X} \rightarrow \mathcal{Y}$ trained on supervised data. Specifically, the markup language we focus on includes latex formulas, HTML codes, musical notations, and chemical molecular sequences. Different from the natural image synthesis benchmarks, the layout of rendered images corresponding to the markups is more discrete, which increases the difficulty of feature extraction. Besides, a rendered image also has a long-term symbol dependency, and thus the accumulation of local errors can lead to information deviation conveyed by the entire image.

\subsection{Input Representation}
As mentioned in section 3.1, the input of markup-to-image is a set of paired markups and rendered images, denoted as $(\textbf{X},\mathbf{Y})$. For a markup sequence $\mathbf{x}$, the pre-trained markup language model \cite{gao2020pile,chithrananda2020chemberta} is employed as an encoder to obtain its embedding representation $\mathbf{t} \in \mathbb{R}^{N \times D}$, where $N$ and $D$ are the number of tokens and the dimension of each token, respectively. Following the previous work \cite{deng2022markup}, we adopt ResNet to extract visual features $\mathbf{v} \in \mathbb{R}^{C \times H \times W}$ for a rendered image $\mathbf{y}$, where $C$, $H$, and $W$ are the channel, height, and width of $\mathbf{v}$, respectively. Since rendered images have a sequential structure with rich contextual information, we propose to progressively optimize visual features and align them with the corresponding markup embeddings to learn robust uni-modal representations. 

\textbf{Sequential Visual Feature Capturing.}
We first use a \emph{convolution + map-to-sequence} operation \cite{aberdam2021sequence} to convert $\mathbf{v}$ into a sequence of visual tokens $\mathbf{v^s}=(\mathbf{v^s}_1,\mathbf{v^s}_2,..., \mathbf{v^s}_M) \in \mathbb{R}^{M \times D}$, where each token corresponds to a receptive field in $\mathbf{y}$ and the number of tokens $M$ depends on the width of $\mathbf{y}$. Afterward, $\mathbf{v^s}$ is fed into a bidirectional LSTM (Bi-LSTM) \cite{li2022unified} to capture the contextual information, allowing each token to distinguish itself from its semantic context as well as capture the long-term symbol dependency. Finally, we refine $\mathbf{v^s}$ by concatenating the hidden states $\mathbf{h}=(\mathbf{h}_1,\mathbf{h}_2,...,\mathbf{h}_M) \in \mathbb{R}^{M \times D}$ output by Bi-LSTM, as suggested in \cite{litman2020scatter}.

\textbf{Fine-grained Cross-modal Alignment.}
Considering that a pair of markup and rendered image inputs sequentially convey consistent semantics, it is important to explicitly mine the sequence-similarity between them. Therefore, we propose a fine-grained alignment pattern at the sequence level as illustrated in Figure \ref{fig:Framework}, aiming at learning more robust uni-modal representations.

Specifically, a cross-attention module $CAM(\cdot,\cdot,\cdot)$ \cite{huang2019ccnet} is first employed to capture the receptive field-to-token relevance between the input image and markup:
\begin{equation}
    \mathbf{c}=CAM(\mathbf{t},\mathbf{v^s},\mathbf{v^s}),
\end{equation}
where $\mathbf{c} \in \mathbb{R}^{N \times D}$ is an evolved feature with the same length as $\mathbf{t}$, and $\mathbf{t}$, $\mathbf{v^s}$ and $\mathbf{v^s}$ are the query, key, and value matrices, respectively. Then, we define a fine-grained alignment loss to optimize the correlation between the two modalities, where $\mathbf{c}_i$ is encouraged to be similar to $\mathbf{t}_i$ with the same index and dissimilar to $\mathbf{t}_j$ with different indexes ($i \neq j$). As cosine similarity $cos(\cdot,\cdot)$ is employed to measure the sequence-similarity, the fine-grained cross-modal alignment loss $\mathcal{L}_{fa}$ can be formulated as:
\begin{equation}\label{fine-grained alignment loss}
    \mathcal{L}_{fa}=\frac{1}{N}\sum^N_{i=1}[1-cos(\mathbf{c}_i, \mathbf{t}_i)+\frac{1}{N-1}\sum^N_{j=1,j \neq i}cos(\mathbf{c}_i \cdot\mathbf{t}_j)].
\end{equation}
The optimization of $\mathcal{L}_{fa}$ allows us to better learn the inter-modal sequence similarity between rendered images and markups, thereby obtaining more robust uni-modal representations.

\subsection{Contrast-augmented Diffusion Model}
Given a variable $y_0$, most of the diffusion models (DMs) maximize the evidence lower bound (ELBO) of $\log{p(y_0)}$ on the Markov chain $q(y_1,...y_T|y_0)=\sum_{t=1}^Tq(y_t|y_{t-1})$:
\begin{equation}
    \begin{aligned}
        \begin{split}
            maximize \log{(y_0)}\geq& \mathbb{E}_q\log{\frac{p(y_{0:T})}{q(y_{1:T}|y_0)}} \\
            =&\mathbb{E}_q[\log{p(y_0|y_1)}-D_{KL}(q(y_T|y_0)||p(y_T))] \\
            &-\sum_{t=1}^T D_{KL}(q(y_{t-1}|y_t,y_0)||p(y_{t-1}|y_t)) \\
            =&\mathcal{L}_{elbo}(y_0),
        \end{split}
    \end{aligned}
\end{equation}
where $q$ denotes the probabilistic distribution of real data, and $p$ is an approximate probability distribution of $q$. $D_{KL}(\cdot||\cdot)$ is Kullback-Leibler Divergence, which is widely used to measure the difference between two distributions. On this basis, our approach introduces contrastive learning into the diffusion model, which exposes the model to both positive and negative samples to improve the generalization ability as well as alleviate the exposure bias problem. Specifically, we explicitly consider the expected states of positive sample $y^{\prime}_0$ and negative sample $\bar{y}_0$ based on the original variational inference, which maximizes the log-likelihood of $y^{\prime}_0$ as well as minimizes the log-likelihood of $\bar{y}_0$, as shown in the following objective:
\begin{equation}\label{our log-likelihood}
    maximize \log{p(y_0,y_0^{\prime})}-\lambda\log{p(\bar{y}_0)},
\end{equation}
where $\lambda$ is a balanced weight. It is very difficult to directly solve Equation \ref{our log-likelihood}, and thus we optimize it to maximize the variational lower bound as follows:
\begin{equation}\label{our elbo-eubo}
    \begin{aligned}
        \begin{split}
            &\log{p(y_0,y_0^{\prime})}-\lambda\log{p(\bar{y}_0)} \\
            \geq&\mathcal{L}_{elbo}(y_0,y_0^{\prime})-\lambda \mathcal{L}_{eubo}(\bar{y}_0),
        \end{split}
    \end{aligned}
\end{equation}
where $eubo$ represents the evidence upper bound. Specifically, we use mild augmentation \cite{aberdam2021sequence} and same-batch sampling strategies \cite{Chen_represent} to create a positive sample $y_0^{\prime}$ and several negative samples $\bar{y}_0$ for each $y_0$, respectively. Below, we will describe the process of solving these two log-likelihood boundaries.

\textbf{ELBO of Positive Log-likelihood.}
Since $y_0^{\prime}$ is another view of $y_0$, we propose to model them in a logarithmic joint likelihood $\log{p(y_0,y_0^{\prime})}$ and take into account $y_t$ generated at each step $t$ during the diffusion process (see Appendix \ref{derivation of elbo term} for details):
\begin{equation}\label{elbo of positive}
    \begin{aligned}
        \begin{split}
            \log{p(y_0,y_0^{\prime})}=&\log \int_{y_1^{\prime},...y_{t-1}^{\prime}} \int_{y_1,...,y_{t-1}}p(y_0,y_t,y_0^{\prime},y_t^{\prime})dy_tdy_t^{\prime} \\
            \geqslant& \mathbb{E}_{q(y_t,y_t^{\prime}|y_0,y_0^{\prime})} \log[\frac{p(y_0,y_t,y_0^{\prime},y_t^{\prime})}{q(y_t,y_t^{\prime}|y_0,y_0^{\prime})}] \\
            =&\mathcal{L}_{elbo}(y_0)+\mathcal{L}_{elbo}(y_0^{\prime})+\mathbb{E}_{y_t,y_t^{\prime}}[MI(y_t,y_t^{\prime})],
        \end{split}
    \end{aligned}
\end{equation}
where $MI(\cdot,\cdot)$ is mutual information \cite{Parmar_2021_CVPR}. From Equation \ref{elbo of positive}, it can be seen that one of the goals is to maximize the mutual information between $y_t$ and $y_t^{\prime}$, which enables the model to emphasize the similarity relationships between samples. By training on these relationships, our model learns to better capture the essential features of similar samples, thereby improving generalization ability.

\textbf{EUBO of Negative Log-likelihood.}
EUBO has favorable properties: it has a mass covering effect advantageous in the approximation of the posterior, and thus provides a tighter bound for the variational evidence. Equation. \ref{our elbo-eubo} can be optimized (see Appendix \ref{derivation of eubo term} for details) by using the method \cite{dieng2017variational}:
\begin{equation}
    \begin{aligned}
        \begin{split}
            \log{p(\bar{y}_0)}&\leq CUBO_{\chi^2}=\frac{1}{2}\log{\mathbb{E}_{q(\bar{y}_t)}[{(\frac{p(\bar{y}_0,\bar{y}_t)}{q(\bar{y}_t)})}^2]} \\
            &\triangleq exp(2CUBO_{\chi^2})=\mathbb{E}_{q(\bar{y}_t|\bar{y}_0)}[{(\frac{p(\bar{y}_0,\bar{y}_t)}{q(\bar{y}_t|\bar{y}_0)})}^2] \\
            &=\mathbb{E}_{q(\bar{y}_t|\bar{y}_0)}(e^{2\log{\frac{p(\bar{y}_0,\bar{y}_t)}{q(\bar{y}_t|\bar{y}_0)}}})=e^{2\mathcal{L}_{elbo}(\bar{y}_0)}.
        \end{split}
    \end{aligned}
\end{equation}

Different from \cite{zhudiscrete} only using negative samples to increase the lower bound, our model explicitly considers the impact of both positive and negative logarithmic likelihoods by placing evidence between tighter upper and lower bounds, resulting in better variational inference for performance improvement.

\textbf{Training.} We jointly train all the components in FSA-CDM, and the final loss function is expressed as:
\begin{equation}\label{final loss function}
    \begin{aligned}
        \begin{split}
    \mathcal{L}_{FSA-CDM}=&\beta \mathcal{L}_{fa}-\mathcal{L}_{elbo}(y_0)-\mathcal{L}_{elbo}(y_0^{\prime}) \\
    &+\lambda e^{2\mathcal{L}_{elbo}(\bar{y}_0)}-\mathbb{E}_{y_t,y_t^{\prime}}[MI(y_t,y_t^{\prime})],
        \end{split}
    \end{aligned}
\end{equation}
where $\beta$ is the weight of fine-grained sequence alignment loss in Equation. \ref{fine-grained alignment loss}. For the  mutual information term, we adopt the contrastive loss $\mathcal{L}_{cl}$ to efficiently approximate it as follows:
\begin{equation}
    \mathcal{L}_{cl}=\log{\frac{\exp(y_t^T \cdot y_t^{\prime} / \tau)}{\sum\exp(y_t^T \cdot \bar{y}_t / \tau)}},
\end{equation}
where $\tau$ is a temperature parameter. As a result, Equation. \ref{final loss function} can be rewritten as:
\begin{equation}
    \mathcal{L}_{FSA-CDM}=\beta \mathcal{L}_{fa}-\mathcal{L}_{elbo}(y_0)-\mathcal{L}_{elbo}(y_0^{\prime})+\lambda e^{2\mathcal{L}_{elbo}(\bar{y}_0)}-\mathcal{L}_{cl}.
\end{equation}

\subsection{Context-aware Cross Attention Module}
\begin{figure}
    \centering
    \includegraphics[width=0.35\textwidth]{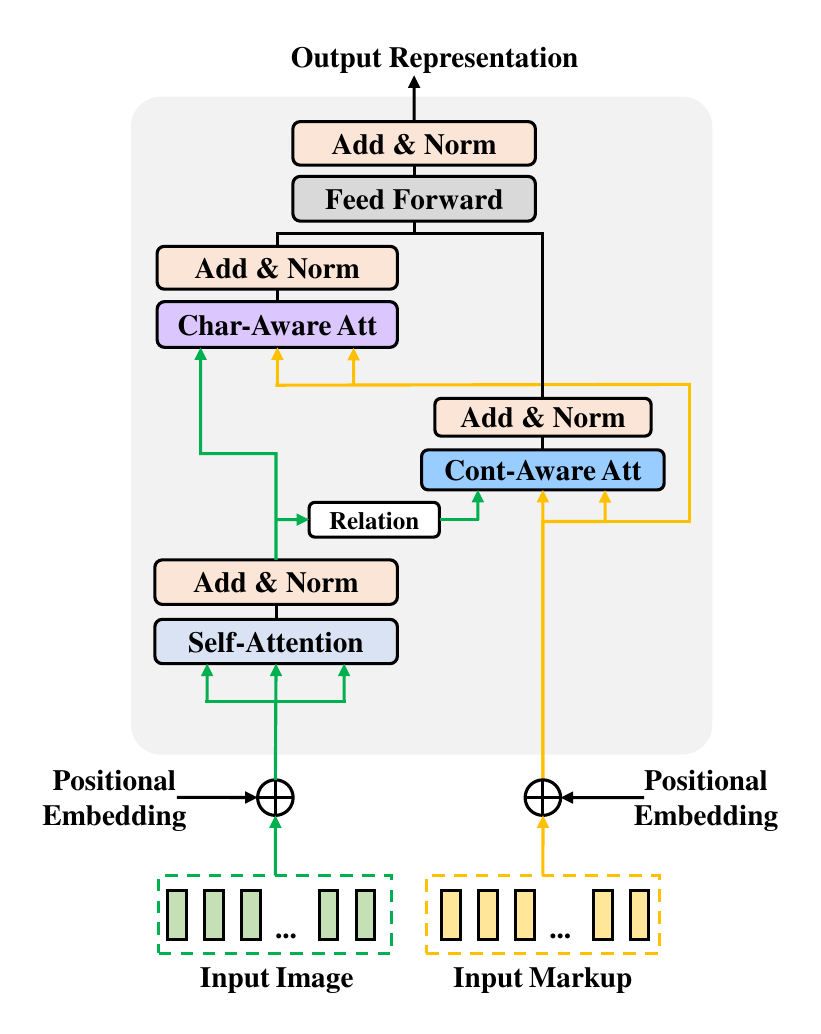}
    \caption{Context-aware Cross Attention Module (CCAM), which consists of a self-attention module, a character-aware attention module, and a context-aware attention module. The green and yellow paths represent visual and markup features, respectively.}
    \label{fig:CCAM}
\end{figure}

The information conveyed by an input markup and its corresponding image mainly includes characters and their contextual correlations. Motivated by this, we design a Context-aware Cross Attention Module (CCAM) to better fuse visual and markup representations during the denoising process.

The structure of CCAM is shown in Figure \ref{fig:CCAM}, which mainly consists of three parts: self-attention (SA), character-aware attention (ChA), and context-aware attention (CoA). The SA is used to capture the internal dependencies between elements of visual features, and its output $\mathbf{v^{sa}}$is fed into parallel ChA and CoA. In ChA, we transform $\mathbf{v^{sa}}$ into the queries $Q_{ca} \in \mathbb{R}^{N \times d_q}$, and markup representations $\mathbf{t}$ into keys $K_{ca} \in \mathbb{R}^{N \times d_k}$ and values $V_{ca} \in \mathbb{R}^{N \times d_v}$. The intuition is that the content of a rendered image is discrete, and thus we need to suppress unrelated parts in $\mathbf{v^{sa}}$ and accurately construct connections between referring characters in $\mathbf{v^{sa}}$ and the corresponding entities in $\mathbf{t}$. Therefore, we perform ChA as:
\begin{equation}
    ChA(Q_{ca},K_{ca},V_{ca})=softmax(\frac{Q_{ca}K_{ca}^T}{\sqrt{d_k}})V_{ca}.
\end{equation}

Besides the exploration of characters, it is also necessary to have a deep understanding of the contextual information formed among characters. For instance, $"\backslash frac"$ is often followed by the numerator and denominator strings in order, which could help the model better understand the contextual semantics. For this purpose, we develop the CoA, which is combined with ChA to finely learn multi-modal features. First, we obtain a relation matrix $\mathcal{R}=r_{ij}$ based on $\mathbf{v^{sa}}$ by using the approach proposed in \cite{cao2021global} with a low computational overhead, where $r_{ij}$ describes the contextual relationship between image region $i$ and region $j$. We linearly project (denoted as $\Psi(\cdot)$) the relation matrix $\mathcal{R}$ as queries, denoted as $Q_{cu} \in \mathbb{R}^{N \times d_q}$:
\begin{equation}
    \mathbf{v} \rightarrow \mathcal{R}=r_{ij},Q_{cu}= \Psi(\mathcal{R}).
\end{equation}
Afterward, we concatenate (denoted as $\Phi(\cdot,\cdot)$) the visual and markup features at the sequence level and send them into two different linear layers to obtain the sequence memory keys $K_{cu} \in \mathbb{R}^{(N+HW) \times d_k}$ and values $V_{cu} \in \mathbb{R}^{(N+HW) \times d_v} $, which contain potential cross-modal semantics: 
\begin{equation}
    K_{cu}= \Psi(\Phi(\mathbf{v},\mathbf{t})), V_{cu}= \Psi(\Phi(\mathbf{v},\mathbf{t})).
\end{equation}
Finally, we take the cross-attention operation to capture the intra- and inter-modality semantic relationship, to help the model better understand contextual information:
\begin{equation}
    CoA(Q_{cu},K_{cu},V_{cu})=softmax(\frac{Q_{cu}K_{cu}^T}{\sqrt{d_k}})V_{cu}.
\end{equation}

Compared to the traditional cross-attention used in the denoising process, our approach could simultaneously capture notable characters and the potential contextual information among them, and thus could better output noises to support image construction. 

\begin{table*}
         \caption{Evaluation results of four advanced approaches on four datasets, where DTW and RMSE are the main evaluation metrics, supplemented by SSIM, PSNR, ERGAS, and RASE. ``Params'' and ``Throughput'' denote the parameter complexity (M) and inference speed (seconds/Img), respectively, where all DMs perform 1000 denoising steps for inference.}\label{comparison with sota methods}
          \resizebox{0.72\textwidth}{!}{
           \begin{tabular}{cccccccccc}
           \toprule       
             \multirow{2}{*}{\centering \textbf{Dataset}} & \multirow{2}{*}{\textbf{Approach}} & \multicolumn{2}{c}{\textbf{Main Metrics}} & \multicolumn{4}{c}{\textbf{Complimentary}} & \multicolumn{2}{c}{\textbf{Inference}}\\
                &  & {\textbf{DTW}↓} & {\textbf{RMSE}↓} & {\textbf{SSIM}↑}& {\textbf{PSNR}↑}& {\textbf{ERGAS}↓}& {\textbf{RASE}↓} & {\textbf{Params}} & {\textbf{Throughput}}\\
            \hline
            \multirow{5}{*}{\textbf{Math}} & XMC-GAN & 20.05 & 38.56 &  0.77  &  16.93  & 2367.48  &  592.55 & 174 & \textless 1   \\
                            & SS-DM & 18.81 & 37.19 &   0.79 &  17.25 & 2247.41 &  561.85 & 209 & 48  \\
                            & CDCD & 17.98  & 36.47 &   0.79 &  17.04 & 2164.14 &  550.31 & 195 & 44 \\
                            & FSA-CDM & \textbf{15.76} & \textbf{34.52} & \textbf{0.81} &  \textbf{18.14} & \textbf{1996.54} &  \textbf{496.23}  & 253 & 53 \\
                              & Improvements & \textbf{$+12.34\%$} & \textbf{$+5.35\%$} & \textbf{$+2.53\%$} &  \textbf{$+6.46\%$} & \textbf{$+7.74\%$} &  \textbf{$+9.83\%$}  & - & - \\
                \hline
                \multirow{5}{*}{\textbf{Simple Tables}} & XMC-GAN & 6.15 & 23.08 &   0.90 &   38.14 & 2523.83  &  657.46  & 174 & \textless 1  \\
                            & SS-DM & 5.64 & 21.11 &   0.93 &  40.20  & 2285.83 &  571.46  & 209 & 46  \\
                            & CDCD & 5.47  & 20.63 &   0.94 &  41.03 &  2176.21 &  557.76 & 195 & 43   \\
                            & FSA-CDM & \textbf{5.03} & \textbf{19.78} & \textbf{0.95} &  \textbf{42.35} &  \textbf{2024.16} &  \textbf{508.14}  & 253 & 51  \\
                             & Improvements & \textbf{$+8.04\%$} & \textbf{$+4.12\%$} & \textbf{$+1.06\%$} &  \textbf{$+3.22\%$} & \textbf{$+6.99\%$} &  \textbf{$+8.90\%$}  & - & - \\
                \hline
                \multirow{5}{*}{\textbf{Sheet Music}} & XMC-GAN & 80.77 & 45.21 &   0.67 &  15.14 & 3036.52 &  761.09  & 174 & \textless 1  \\
                            & SS-DM & 79.76 & 44.70 & 0.68 &  15.20 & 2978.36 &  744.59  & 209 & 137  \\
                            & CDCD &  78.93 & 44.26 & 0.69 &  15.35 & 2937.57 &  733.05 & 195 &  127  \\
                            & FSA-CDM & \textbf{76.79} & \textbf{43.41} & \textbf{0.71} & \textbf{15.73} & \textbf{2866.71} & \textbf{707.24} & 253 & 151 \\
                            & Improvements & \textbf{$+2.71\%$} & \textbf{$+1.92\%$} & \textbf{$+2.90\%$} &  \textbf{$+2.48\%$} & \textbf{$+2.41\%$} &  \textbf{$+3.52\%$} & - & -  \\
                \hline
                \multirow{5}{*}{\textbf{Molecules}} & XMC-GAN & 25.04 & 38.22 &  0.60  &  16.60 & 2496.33 &  623.58 & 174 & \textless 1   \\
                            & SS-DM & 24.80 & 37.92 &  0.61 &  16.69 & 2467.16 &  616.79 & 209 & 126   \\
                            & CDCD & 24.31 & 36.86 & 0.63 & 16.87 & 2415.08 & 600.32  & 195 & 117  \\
                            & FSA-CDM & \textbf{23.69} & \textbf{36.14}&  \textbf{0.63} &  \textbf{17.06} & \textbf{2386.35} &  \textbf{574.57} & 253 & 139  \\
                            & Improvements & \textbf{$+2.55\%$} & \textbf{$+1.95\%$} & \textbf{$+0.00\%$} &  \textbf{$+1.12\%$} & \textbf{$+1.19\%$} &  \textbf{$+4.29\%$} & - & -  \\
                \bottomrule
    \end{tabular}
}
\end{table*}

\section{Experiments} \label{experiments}
\subsection{Datasets}
We conduct experiments on datasets from four domains:

\textbf{Math} is a large collection of real-world mathematical expressions written in LaTeX markups and their rendered images. There are a total of $55,033$ training, $6,072$ validation, and  $1,024$ testing text-image pairs, where the image size is $64 \times 320$ and the input markup contains $113$ characters on average.

\textbf{Simple Tables} were collected based on the 100k synthesized HTML snippets and the corresponding rendered webpage images. There are $80,000$ training, $10,000$ validation, and $1,024$ testing pairs, where the image size is $64 \times 64$ with an average of $481$ characters in markup text.

\textbf{Sheet Music} adopts LilyPond files as its markup language, and generates $32,880$ markup-image pairs, including $30,902$ for training, $989$ for validation, and $988$ for testing. The image size is $192 \times 448$ with an average of 240 characters in markup text.

\textbf{Molecule} from the chemistry domain contains $19,925$ 2D molecules images specified by SMILES strings. It is divided into $17,925$ training, $1,000$ validation, and $1,000$ testing samples. Different from the other three datasets, the rendered image in Molecules is colored with a size of $128 \times 128$ and an average length of $30$ in markup text.

\subsection{Implementation Details}
\textbf{Experimental Settings.} Following \cite{deng2022markup}, we initiate the markup encoder with GPT-Neo-175M \cite{gao2020pile} for Math, Simple Tables, and Sheet Music datasets and ChemBert-77M-MLM \cite{chithrananda2020chemberta} for Molecules dataset. For input images, we build a lightweight ResNet as the image encoder. The relevant settings of our diffusion scheduler refer to the HuggingFACE diffusers library\footnote{https://github.com/huggingface/diffusers}, where the noise estimator is a U-Net with five CCAMs and eight conventional cross-attention layers. We use mild augmentation strategy \cite{aberdam2021sequence} to create a positive sample that maintains semantic consistency for each training sample. And five negative samples are sampled from the same batch \cite{Chen_represent}, which have different semantics from the positive sample.
The weights $\lambda$ and $\beta$ are set to 0.005 and 0.02, respectively. We set the batch sizes for Math, Simple Tables, Music and Molecules are 16, 24, 8 and 16, respectively, and we train all models for 100 epochs using the Adam optimizer with the learning rate $0.0001$ on 4 Nvidia RTX A6000 with 48 GPU VRAM. The learning rate is decreased by the cosine decay strategy with 500 warmup steps.

\textbf{Metrics.} For the markup-to-image task, the generated image should be consistent with the ground truth image at the pixel level, which is different from other popular image-generation tasks. Therefore, we adopt Dynamic Time Warping (DTW) and Root Squared Mean Error (RMSE) as our main evaluation metrics following \cite{deng2022markup}. RMSE compares two images at the pixel level, and it penalizes the generated image with smaller character offsets, even if it is semantically equivalent to the ground truth image. DTW calculates pixel-level similarity by treating an image as time series through binarization and allows minor offsets of the generated image, which fits the markup-to-image task. Structural similarity index measure (SSIM), peak signal-to-noise ratio (PSNR), erreur relative globale adimensionnelle de synthèse (ERGAS), and relative average spectral error (RASE) serve as supplements for the above two main metrics to report performance more comprehensively.

\subsection{Comparison with state-of-the-art methods}
We compare the proposed method with several SOTA methods, including XMC-GAN~\cite{Zhanghan_2021_cvpr}, SS-DM~\cite{deng2022markup}, and CDCD~\cite{zhudiscrete} on Math, Simple Tables, Sheet Music, and Molecules datasets. All the approaches except SS-DM are replicated by using the optimization parameter settings.

\textbf{Quantitative Analysis.} We calculate six evaluation metrics mentioned above and show the results in Table \ref{comparison with sota methods}. Benefitting from the stable training of the diffusion model, SS-DM achieves better performance as compared to XMC-GAN, and CDCD further improves the performance due to the introduction of negative samples by using contrastive learning during the denoising process. Comparatively, our FSA-CDM achieves the best performance with a significant improvement on all the metrics because FSA-CDM introduces a robust feature learning module and contrastive positive and negative samples for the diffusion model, generating stronger generalization ability. In addition, we discover that the DTW improvement by FSA-CDM is greater than the RMSE improvement. This is because DTW is a relaxed pixel-level similarity and is consistent with our encoding process, which divides a text-image pair into a sequence of visual and textual tokens with the alignment on the token level. What is more, the improvements in the Math and Tables domains are more significant than that in Music and Molecules since Music and Molecules have a long dependence chain of symbols, which is too difficult to be captured. Despite the success, DM-based approaches are much slower than GAN-based approaches like XMC-GAN since they require multiple denoising steps. Our FSA-CDM is inferior to CDCD and SS-DM in terms of parameters and inference speed. The increase in model parameters stems from the feature extraction module and CCAM, and the inference time is mainly affected by CCAM during the denoising process. The introduction of contrastive samples has no effect on the inference time. Fortunately, the delay is tolerable and can be alleviated by reducing the denoising steps at the expense of accuracy \cite{Gu_VQ-diffusion}.

\begin{figure*}
    \centering
    \includegraphics[width=0.82\textwidth]{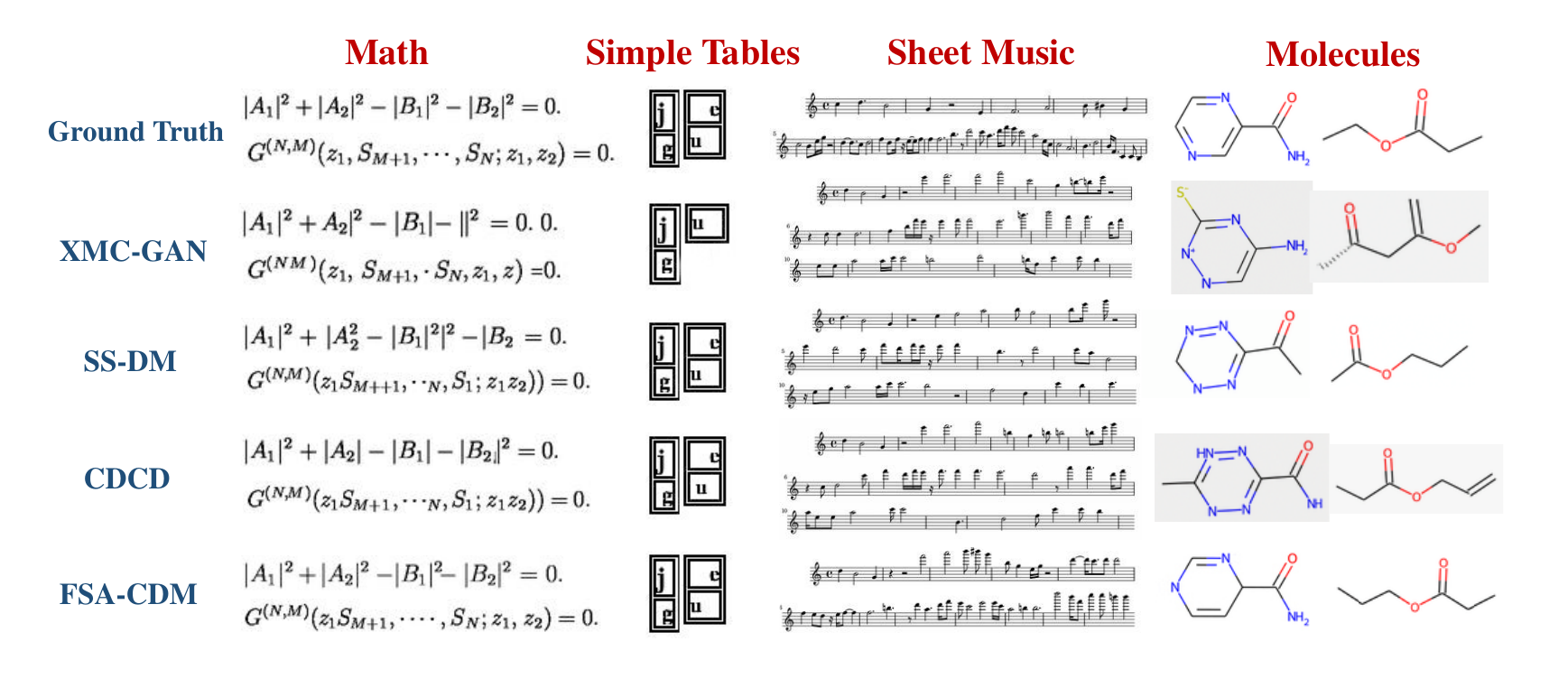}
    \caption{Qualitative results on four datasets. The columns from left to right are Math and Simple Tables, Sheet Music, and Molecules, respectively. And the rows from top to bottom are ground truth, XMC-GAN, SS-DM, CDCD, and FSA-CDM, respectively.}
    \label{fig:qualitative analysis}
\end{figure*}

\textbf{Qualitative Analysis.} Figure \ref{fig:qualitative analysis} presents several markup images generated by SOTA methods and the corresponding ground truth, qualitatively comparing the performance of markup-to-image generation. All the methods exhibit different adaptations to the four domains with varying degrees of difficulty. Specifically, these methods perform well in math and simple tables domain, and our FSA-CDM achieves the best visual effects for human eyes. For the music domain, due to the bottleneck of the long dependence chain of symbols from left to right, it is difficult for the model to maintain accuracy after generating the top few symbols due to the limited number of denoising steps. The molecules image generated by FSA-CDM is the closest to the ground truth, considering the non-uniqueness of the molecules' layout and orientation.

\subsection{Ablation Study} This experiment verifies the effectiveness of the proposed three components, including the fine-grained sequence module (\emph{FSA.}), the contrast-augmented diffusion model (\emph{Con-aug.}), and the context-aware cross attention module (\emph{CCAM.}). Table \ref{ablation study} demonstrates the comparison results on four benchmarks, where we remove the components of \emph{FSA.}, \emph{Con-aug.} and \emph{CCAM.} as our baseline model. From the table, we can draw the following conclusions: First, the introduction of \emph{FSA.} decreases DTW by $3.03\%$, $2.59\%$, $1.47\%$, and $1.03\%$ than the baseline model on Math, Simple Tables, Sheet Music, and Molecules, respectively, which indicates that the sequence alignment capturing between markups and images by our encoding is effective, yielding robust uni-modal representations at a fine-grained level. Second, compared with the baseline network, we explicitly introduce contrastive positive and negative samples in the denoising process, bringing $12.02\%$, $8.80\%$, $2.73\%$, and $2.93\%$ DTW improvements on four benchmarks, respectively. This superior performance gain proves that contrastive samples can encourage the diffusion model to capture more similar and discriminative information during training to improve its generalization ability. Moreover, when \emph{Con-aug.} is combined with \emph{FSA.}, it can achieve an average DTW improvement of $8.84\%$ compared to the baseline model. Third, benefiting from a deeper exploration of character and contextual correlations, the proposed CCAM achieves $6.95\%$, $6.21\%$, $2.05\%$, and $1.77\%$ DTW improvements, respectively. Finally, our approach combining the three proposed components decreases the average DTW and RMSE by $10.21\%$ and $6.35\%$, which is significantly better than the baseline model. 

\begin{table}
         \caption{Evaluation results of ablation studies on four datasets.}\label{ablation study}
          \resizebox{0.45\textwidth}{!}{
           \begin{tabular}{cccccc}
           \toprule       
             \multirow{2}{*}{\centering \textbf{Dataset}} & \multicolumn{3}{c}{\textbf{Different Setting}} & \multicolumn{2}{c}{\textbf{Main Metrics}} \\
                & \emph{FSA.} & \emph{Con-aug.} & \emph{CCAM} & {\textbf{DTW}↓} & {\textbf{RMSE}↓}  \\
            \hline
            \multirow{6}{*}{\textbf{Math}} & - & - & - &  19.12 &  37.54 \\
                            & \checkmark & - & - &  18.54 &  36.85  \\
                            & - & \checkmark & - &  16.82 &  35.47  \\
                            & - & - & \checkmark &  17.79 &  36.44  \\
                            & \checkmark & \checkmark & - & 16.23  &  34.96  \\
                            & \checkmark & \checkmark & \checkmark &  \textbf{15.76} & \textbf{34.52}  \\
                \hline
            \multirow{6}{*}{\textbf{Simple Tables}} & - & - & - &  5.79 &  21.34 \\
                            & \checkmark & - & - &  5.64 &  20.96  \\
                            & - & \checkmark & - &  5.28 &  20.29  \\
                            & - & - & \checkmark &  5.43 &  20.67  \\
                            & \checkmark & \checkmark & - &  5.11 &  19.96  \\
                            & \checkmark & \checkmark & \checkmark &  \textbf{5.03} & \textbf{19.78}  \\
                \hline
            \multirow{6}{*}{\textbf{Sheet Music}} & - & - & - &  79.85 &  44.92 \\
                            & \checkmark & - & - &  78.68 &  44.35  \\
                            & - & \checkmark & - &  77.67 &  43.79  \\
                            & - & - & \checkmark &  78.21 &  44.07  \\
                            & \checkmark & \checkmark & - &  77.17 &  43.58  \\
                            & \checkmark & \checkmark & \checkmark &  \textbf{76.79} & \textbf{43.41}  \\
                \hline
            \multirow{6}{*}{\textbf{Molecules}} & - & - & - &  24.91 &  38.05 \\
                            & \checkmark & - & - &  24.67 &  37.64  \\
                            & - & \checkmark & - &  24.18 &  36.83  \\
                            & - & - & \checkmark &  24.47 &  37.26  \\
                            & \checkmark & \checkmark & - &  23.85 &  36.42  \\
                            & \checkmark & \checkmark & \checkmark &  \textbf{23.69} & \textbf{36.14}  \\
                \bottomrule
    \end{tabular}
}
\end{table}

\subsection{Analysis of Contrastive Variational Loss}
The balanced weight $\lambda$ in contrastive variational loss is a sensitive parameter. To show how $\lambda$ affects the performance of markup-to-image generation, we set different values of $\lambda$ to observe the performance change on Math and Molecules datasets as shown in Table \ref{balanced weight}. When $\lambda$ is equal to $0$, \textit{i.e.} ignoring the optimization of the variational evidence bound by negative samples, we can observe that the performance is the worst. As $\lambda$ increases from $0$ to $0.005$, the performance is consistently improved, which proves the effectiveness of negative variational loss since negative samples can provide dissimilar information for our model to learn differential features between positive and negative samples, thereby improving generalization ability. However, the further increase of $\lambda$ would lead to a performance drop because a too-large value of $\lambda$ would overshadow the utility of positive samples, which are the dominant elements in the denoising process.  

\begin{table}
         \caption{Evaluation results of contrastive variational loss with different weights $\lambda$ on Math and Molecules datasets.}\label{balanced weight}
          \resizebox{0.45\textwidth}{!}
          {
           \begin{tabular}{cccccc}
           \toprule       
             $\lambda$ & 0 & 0.002 & 0.005 & 0.01 & 0.02 \\
             \hline 
             \textbf{DTW on Math} & 16.34  & 16.07 & \textbf{15.76} & 15.94 & 16.29 \\
             \textbf{RMSE on Math} & 35.18 & 34.90 &  \textbf{34.52} & 34.77 & 35.05 \\
             \textbf{DTW on Molecules} & 24.38  &  24.02 & \textbf{23.69} & 23.76 & 24.06 \\
             \textbf{RMSE on Molecules} & 37.21 &  36.65 & \textbf{36.14} & 36.32 & 36.68 \\
             \bottomrule
            \end{tabular}
}
\end{table}

\section{Conclusions} \label{Conclusions}
In this work, we propose a novel ``Contrast-augmented Diffusion Model with Fine-grained Sequence Alignment’’ (FSA-CDM) for markup-to-image generation. Beyond the existing diffusion models, our approach first incorporates positive and negative samples with a novel contrastive variational objective to offer a tighter bound to improve the model’s generalization ability, which generates important theoretical values. In addition, different from natural image generation, our approach considers the characteristics of markup images, including the sequence correlations by using the fine-grained sequence alignment and the complex context by designing the context-aware cross attention module. Extensive experiments on four benchmark datasets confirm the effectiveness of the proposed approach. However, it still has limitations in terms of long dependency chains and inference speed. In future work, we will further explore these issues and attempt to apply the proposed theory to other generation tasks.

\begin{acks}
This work was supported by the National Natural Science Foundation of China (No. 62272157, No. U21A20518, and No. 61976086) and the Natural Science Foundation of Changsha (No. kq2202177).
\end{acks}

\bibliographystyle{ACM-Reference-Format}
\bibliography{sample-base}

\appendix
\section{Appendix}
\subsection{Derivation of ELBO Term}\label{derivation of elbo term}
Given the normal sample $y_0$, positive sample $y_0^{\prime}$, and the intermediate variables $\{y_t\}_{t=1}^T$ and $\{y_t^{\prime}\}_{t=1}^T$ from the diffusion process, we present the full derivation of the evidence lower bound of them:
\begin{equation}
    \begin{aligned}
        \begin{split}
            \log{p(y_0,y_0^{\prime})}=&\log \int_{y_1^{\prime},...y_{t-1}^{\prime}} \int_{y_1,...,y_{t-1}}p(y_0,y_t,y_0^{\prime},y_t^{\prime})dy_tdy_t^{\prime} \\
            =&\log \mathbb{E}_{q(y_t,y_t^{\prime}|y_0,y_0^{\prime})}[\frac{p(y_0,y_t,y_0^{\prime},y_t^{\prime})}{q(y_t,y_t^{\prime}|y_0,y_0^{\prime})}] \\
            \geqslant& \mathbb{E}_{q(y_t,y_t^{\prime}|y_0,y_0^{\prime})} \log[\frac{p(y_0,y_t,y_0^{\prime},y_t^{\prime})}{q(y_t,y_t^{\prime}|y_0,y_0^{\prime})}] \\
            =&\mathbb{E}_{q(y_t,y_t^{\prime}|y_0,y_0^{\prime})} \log[\frac{p(y_0|y_t)p(y_0^{\prime}|y_t^{\prime})p(y_t,y_t^{\prime})}{q(y_t|y_0)q(y_t^{\prime}|y_0^{\prime})}] \\            =&\mathbb{E}_{q(y_t|y_0)}\log[p(y_0|y_t)]+\mathbb{E}_{q(y_t^{\prime}|y_0^{\prime})}\log[p(y_0^{\prime}|y_t^{\prime})] \\
            &+\mathbb{E}_{q(y_t,y_t^{\prime}|y_0,y_0^{\prime})}\log[\frac{p(y_t,y_t^{\prime})}{q(y_t|y_0)q(y_t^{\prime}|y_0^{\prime})}] \\
            =&\mathbb{E}_{q(y_t|y_0)}\log[p(y_0|y_t)]+\mathbb{E}_{q(y_t^{\prime}|y_0^{\prime})}\log[p(y_0^{\prime}|y_t^{\prime})] \\
            &+\mathbb{E}_{q(y_t,y_t^{\prime}|y_0,y_0^{\prime})}\log[\frac{p(y_t,y_t^{\prime})}{p(y_t)p(y_t^{\prime})}] \\
            &+\mathbb{E}_{q(y_t,y_t^{\prime}|y_0,y_0^{\prime})}\log[\frac{p(y_t)p(y_t^{\prime})}{p(y_t|y_0)p(y_t^{\prime}|y_0^{\prime})}] \\       =&\mathbb{E}_{q(y_t|y_0)}\log[p(y_0|y_t)]+\mathbb{E}_{q(y_t^{\prime}|y_0^{\prime})}\log[p(y_0^{\prime}|y_t^{\prime})] \\
            &+\mathbb{E}_{q(y_t,y_t^{\prime}|y_0,y_0^{\prime})}\log[\frac{p(y_t,y_t^{\prime})}{p(y_t)p(y_t^{\prime})}] \\
            &-D_{KL}[q(y_t|y_0)||p(y_t)]-D_{KL}[q(y_t^{\prime}|y_0^{\prime})||p(y_t^{\prime})] \\
            =&\mathbb{E}_{q(y_1|y_0)}\log{p(y_0|y_1)}+\mathbb{E}_{q(y_1^{\prime}|y_0^{\prime})}\log{p(y_0^{\prime}|y_1^{\prime})} \\
            &-\sum_{t=1}^T D_{KL}(q(y_{t-1}|y_t,y_0)||p(y_{t-1}|y_t)) \\
            &-\sum_{t=1}^T{D_{KL}(q(y_{t-1}^{\prime}|y_t^{\prime},y_0^{\prime})||p(y_{t-1}^{\prime}|y_t^{\prime}))} \\
            &-D_{KL}[q(y_t|y_0)||p(y_t)]-D_{KL}[q(y_t^{\prime}|y_0^{\prime})||p(y_t^{\prime})] \\
            &+\mathbb{E}_{q(y_t,y_t^{\prime})}\log[\frac{p(y_t,y_t^{\prime})}{p(y_t)p(y_t^{\prime})}].
        \end{split}
    \end{aligned}
\end{equation}
It is hard to directly solve $\mathbf{E}_{q(y_t,y_t^{\prime})}\log[\frac{p(y_t,y_t^{\prime})}{p(y_t)p(y_t^{\prime})}]$, and here we assume that $p(y_t,y_t^{\prime})=q(y_t,y_t^{\prime})$, $p(y_t)=q(y_t)$ and $p(y_t^{\prime})=q(y_t^{\prime})$ as suggested by \cite{wang2022contrastvae}, then the last term can be written as follows:
\begin{equation}
    \begin{aligned}
        \begin{split}
            \log{p(y_0,y_0^{\prime})}&\geq \mathbf{E}_{q(y_1|y_0)}\log{p(y_0|y_1)}+\mathbf{E}_{q(y_1^{\prime}|y_0^{\prime})}\log{p(y_0^{\prime}|y_1^{\prime})} \\
            &-\sum_{t=1}^T D_{KL}(q(y_{t-1}|y_t,y_0)||p(y_{t-1}|y_t)) \\
            &-\sum_{t=1}^T{D_{KL}(q(y_{t-1}^{\prime}|y_t^{\prime},y_0^{\prime})||p(y_{t-1}^{\prime}|y_t^{\prime}))} \\
            &-D_{KL}[q(y_t|y_0)||p(y_t)]-D_{KL}[q(y_t^{\prime}|y_0^{\prime})||p(y_t^{\prime})]\\
            &+MI(y_t,y_t^{\prime})
        \end{split}
    \end{aligned}
\end{equation}

\subsection{Derivation of EUBO Term}\label{derivation of eubo term}
Given the negative sample $\bar{y}_0$ and the intermediate variables $\{\bar{y}_t\}_{t=1}^T$ from the diffusion process, we present the full derivation of the evidence upper bound of them:
\begin{equation}
    \begin{aligned}
        \begin{split}
            \log{p(\bar{y}_0)}\leq& CUBO_{\chi^2}=\frac{1}{2}\log{\mathbb{E}_{q(\bar{y}_t)}[{(\frac{p(\bar{y}_0,\bar{y}_t)}{q(\bar{y}_t)})}^2]} \\
            \triangleq& exp(2CUBO_{\chi^2})=\mathbb{E}_{q(\bar{y}_t|\bar{y}_0)}[{(\frac{p(\bar{y}_0,\bar{y}_t)}{q(\bar{y}_t|\bar{y}_0)})}^2] \\
            =&\mathbb{E}_{q(\bar{y}_{1:T}|\bar{y}_0)}exp[\log{(\frac{p(\bar{y}_{0:T})}{q(\bar{y}_{1:T}|\bar{y}_0)})}^2] \\
            =&\mathbb{E}_{q(\bar{y}_{1:T}|\bar{y}_0)}exp[2\log{\frac{p(\bar{y}_{0:T})}{q(\bar{y}_{1:T}|\bar{y}_0)}}] \\
            =&\mathbb{E}_{q(\bar{y}_{1:T}|\bar{y}_0)}exp[2\log{\frac{p(\bar{y}_T) \prod_{t=1}^Tp(\bar{y}_{t-1}|\bar{y}_t)}{\prod_{t=1}^Tq(\bar{y}_{t}|\bar{y}_{t-1})}}] \\
            =&\mathbb{E}_{q(\bar{y}_{1:T}|\bar{y}_0)}exp[2\log{\frac{p(\bar{y}_T)p(\bar{y}_0|\bar{y}_1) \prod_{t=2}^Tp(\bar{y}_{t-1}|\bar{y}_t)}{q(\bar{y}_T|\bar{y}_{T-1})\prod_{t=1}^{T-1}q(\bar{y}_{t}|\bar{y}_{t-1})}}] \\
            =&\mathbb{E}_{q(\bar{y}_{1:T}|\bar{y}_0)}exp[2\log{\frac{p(\bar{y}_T)p(\bar{y}_0|\bar{y}_1)}{q(\bar{y}_T|\bar{y}_{T-1})}}+2\log{\prod_{t=1}^{T-1}\frac{p(\bar{y}_t|\bar{y}_{t+1})}{q(\bar{y}_{t}|\bar{y}_{t-1})}}] \\
            =&exp[\mathbb{E}_{q(\bar{y}_1|\bar{y}_0)}2\log{p(\bar{y}_0|\bar{y}_1)}+\mathbb{E}_{q(\bar{y}_{T-1},\bar{y}_T|\bar{y}_0)}2\log{\frac{p(\bar{y}_T)}{q(\bar{y}_T|\bar{y}_{T-1})}} \\
            &+\sum_{t=1}^{T-1}\mathbb{E}_{q(\bar{y}_{1:T}|\bar{y}_0)}\log{\frac{p(\bar{y}_T)}{q(\bar{y}_T|\bar{y}_{T-1})}}] \\
            &-\sum_{t=1}^{T-1}\mathbb{E}_{q(\bar{y}_{1:T}|\bar{y}_0)}2D_{KL}(q(\bar{y}_t|\bar{y}_{t-1})||p(\bar{y}_t|\bar{y}_{t+1}))] \\
            =&exp[\mathbb{E}_{q(\bar{y}_1|\bar{y}_0)}2\log{p(\bar{y}_0|\bar{y}_1)} \\
            &-\mathbb{E}_{q(\bar{y}_{T-1}|\bar{y}_0)}2D_{KL}(q(\bar{y}_T|\bar{y}_{T-1})||p(\bar{y}_T))\\
            =&e^{2\mathcal{L}_{elbo}(\bar{y}_0)}.
        \end{split}
    \end{aligned}
\end{equation}

\end{document}